\pdfoutput=1

\documentclass[11pt]{article}

\usepackage[final]{acl}

\usepackage{times}
\usepackage{latexsym}
\usepackage[T1]{fontenc}
\usepackage[utf8]{inputenc}
\usepackage{microtype}
\PassOptionsToPackage{hyphens}{url} 
\usepackage{url}                    
\usepackage{ragged2e,array,tabularx,placeins}
\newcolumntype{L}[1]{>{\RaggedRight\arraybackslash}p{#1}} 
\usepackage{graphicx}
\usepackage{booktabs}
\usepackage{amsfonts}
\usepackage{nicefrac}
\usepackage{colortbl}
\usepackage{listings}
\usepackage{textcomp}
\usepackage{newunicodechar}
\usepackage{hyperref}

\newunicodechar{ọ}{\d{o}}
\newunicodechar{ẹ}{\d{e}}

\title{Yankari: Monolingual Yoruba Dataset}
\author{
  Maro Akpobi \\
  African Center for Language Preservation \\
  \texttt{maro@acflp.org}\\
}

\begin{document}
\maketitle

\begin{abstract}
This paper presents Yankari, a large-scale monolingual dataset for the Yoruba language, aimed at addressing the critical gap in Natural Language Processing (NLP) resources for this important West African language. Despite being spoken by over 30 million people, Yoruba has been severely underrepresented in NLP research and applications. We detail our methodology for creating this dataset, which includes careful source selection, automated quality control, and rigorous data cleaning processes. The Yankari dataset comprises 51,407 documents from 13 diverse sources, totaling over 30 million tokens. Our approach focuses on ethical data collection practices, avoiding problematic sources and addressing issues prevalent in existing datasets. We provide thorough automated evaluations of the dataset, demonstrating its quality compared to existing resources. The Yankari dataset represents a significant advancement in Yoruba language resources, providing a foundation for developing more accurate NLP models, supporting comparative linguistic studies, and contributing to the digital accessibility of the Yoruba language.
\end{abstract}

\section{Introduction}
Natural Language Processing (NLP) has made tremendous strides in recent years, yet these advancements have primarily benefited high-resource languages, leaving many African languages, including Yoruba, underrepresented in NLP research and applications. This paper introduces Yankari, a large-scale, high-quality monolingual dataset for Yoruba, a language spoken by over 30 million people in West Africa. Despite its significant speaker population, Yoruba has long suffered from a lack of comprehensive, ethically-sourced language resources suitable for modern NLP tasks.

Yankari addresses this critical gap by providing a carefully curated corpus of 51,407 documents from 13 diverse sources, totaling over 30 million tokens. Our methodology prioritizes ethical data collection, rigorous quality control, and the preservation of linguistic authenticity. By avoiding problematic sources such as religious texts and machine-translated content, Yankari offers a more balanced and representative sample of contemporary Yoruba language use.

This paper details our data collection and processing pipeline, discusses the challenges encountered in creating resources for low-resource languages, and provides a thorough analysis of the dataset's composition and potential biases. We also offer insights into the ethical considerations surrounding the creation and use of such resources. Through Yankari, we aim to facilitate the development of more accurate and culturally appropriate NLP models for Yoruba, contribute to the preservation of linguistic diversity in the digital age, and provide a replicable approach for creating high-quality datasets for other low-resource languages.

To our knowledge, Yankari represents the first large-scale, non-religious domain monolingual resource created specifically for Yoruba. This work not only provides a valuable asset for Yoruba NLP but also offers a replicable approach for developing similar datasets for other low-resource languages.
\begin{table*}[t]
\small
\centering
\begin{tabularx}{\textwidth}{@{}L{4cm} r L{2.8cm} L{3cm}@{}}
\toprule
\textbf{Source} & \textbf{Docs} & \textbf{Type} & \textbf{Domain} \\
\midrule
\url{yo.wikipedia.org}             & 16\,809 & Encyclopedia & General \\
\url{alaroye.org}                  & 10\,535 & News         & Current Affairs \\
\url{www.bbc.com}                  & 8\,252  & News         & Current Affairs \\
\url{www.awikonko.com.ng}          & 5\,438  & Blog         & Culture \\
\url{yoruba.von.gov.ng}            & 2\,542  & News         & Current Affairs \\
\url{sportsinyoruba.wordpress.com} & 2\,328  & Blog         & Sports \\
\url{www.asejere.net}              & 2\,079  & Blog         & Entertainment \\
\url{asa.ooduarere.com}            & 1\,744  & Blog         & Culture \\
\url{radionigeriaibadan.gov.ng}    &   824  & News         & Current Affairs \\
\url{iroyinowuro.com.ng}           &   603  & News         & Current Affairs \\
\url{oroyoruba.blogspot.com}       &   139  & Blog         & Culture \\
\url{yo.globalvoices.org}          &    81  & News         & Current Affairs \\
\url{edeyorubarewa.com}            &    33  & Blog         & Fashion \\
\midrule
\textbf{Total}                     & 51\,407 &              &         \\
\bottomrule
\end{tabularx}
\caption{Data sources for the Yankari dataset.}
\label{table:datasources}
\end{table*}
\FloatBarrier            

\section{Related Works}
This section examines existing resources for Yoruba NLP, highlighting their limitations and the need for a comprehensive, ethically-sourced Yoruba dataset.

\subsection{Monolingual Yoruba Datasets}
\subsubsection{Yorùbá Text C3}
\citet{alabi2020massive} introduced Yorùbá Text C3, compiled from various web sources including the Bible, JW300 \citep{agic-vulic-2019-jw300}, books, news articles, and Wikipedia. While broad in scope, this dataset is heavily skewed towards religious content, particularly Christianity. This bias significantly limits its utility for NLG tasks requiring balanced and diverse text. Moreover, the inclusion of JW300 data raises serious ethical and legal concerns. \citet{hutchinson2024modeling} points out that the Jehovah's Witnesses have explicitly prohibited the use of their data in NLP research, making the continued use of JW300 not just ethically questionable but potentially illegal.

\subsubsection{MENYO-20k}
\citet{adelani2021menyo} developed MENYO-20k, a multi-domain English-Yoruba corpus primarily for machine translation tasks. While it offers more diverse content, its relatively small size of 20,100 sentences and focus on translation rather than monolingual text generation limit its applicability for large-scale NLG tasks.

\subsection{Multilingual Datasets Including Yoruba}
\subsubsection{Wura Dataset}
The Wura dataset, developed by \citet{oladipo2023wura}, is a multilingual dataset containing approximately 68,000 Yoruba documents. It integrates content from JW300 and Wikipedia, inheriting similar biases and ethical issues as Yorùbá Text C3. Our manual inspection of the Wura dataset revealed critical quality issues not previously reported, including formatting errors and inappropriate content.

\subsubsection{Large-Scale Web-Crawled Corpora}
Multilingual datasets such as mC4, OSCAR, and the Afriberta-Corpus also include Yoruba content. However, these datasets often suffer from noise, poor formatting, and limited source diversity. \citet{ogueji2021small} used the Afriberta-Corpus, which primarily sources data from the BBC News and Common Crawl, resulting in a dataset lacking the domain diversity necessary for robust NLG.

\citet{nguyen2023culturax} introduced CulturaX, covering 167 languages. However, its Yoruba subset showed an alarmingly high duplication rate of 24.48\% and contained machine-translated pages, false positives in language detection, and a significant amount of religious text.

\subsection{Large-Scale Multilingual Efforts for African Languages}
The Cheetah project by \citet{adebara2024cheetahnaturallanguagegeneration} focuses on natural language generation for 517 African languages, including Yoruba. This ambitious project utilizes existing corpora from various sources, including OSCAR \citep{OrtizSuarez2020}, CC-100 \citep{Conneau2020}, Afriberta-Corpus \citep{ogueji2021small}, and mC4 \citep{xue2021mt5}.

While Cheetah represents a significant step towards improving NLP capabilities for African languages, it faces challenges common to large-scale multilingual efforts. These include potential issues with data quality, bias towards certain domains or text types, and the inclusion of machine-translated content.

\subsection{Ethical Considerations in Using Religious Texts}
The ethical implications of using religious texts in NLP, particularly for low-resource languages, are profound and often overlooked. \citet{hutchinson2024modeling} challenges the NLP community's casual approach to sacred texts, arguing that the prevalence of Christian texts in datasets for low-resource languages reflects a legacy of colonialism and missionary work. This creates an ethical dilemma where NLP technologies risk becoming unwitting agents of cultural imperialism and religious proselytism.

\section{Motivation for Yankari}
Given these severe limitations and ethical concerns in existing resources, there is an urgent need for a high-quality, diverse, and extensive monolingual Yoruba dataset that does not rely on restricted or problematic sources. This is where \textbf{Yankari} comes in, directly addressing the gaps and ethical issues left by previous datasets.

Yankari aims to provide a large-scale, ethically sourced corpus that represents a wide range of Yoruba language use. By avoiding the pitfalls of previous datasets, such as over-reliance on religious texts or machine-translated content, Yankari seeks to offer a more balanced and authentic representation of the Yoruba language. This approach aligns with recent calls in the NLP community for more thoughtful and ethical dataset creation, particularly for low-resource languages.

\section{Methodology}
This section details our approach to creating the Yankari dataset, including data collection, processing, quality assurance steps, and corpus analysis.

\subsection{Data Collection}
Our data collection process focused on gathering content from diverse, high-quality sources to ensure a representative sample of contemporary Yoruba language use. We carefully selected 13 sources, including news outlets, blogs, educational websites, and Wikipedia. Table \ref{table:datasources} provides an overview of these sources and their contributions to the dataset.

\subsection{Analysis of Existing Datasets}
To inform our data collection and curation process, we conducted a detailed analysis of the Wura dataset \citep{oladipo2023wura}. Our investigation revealed several critical issues:

\begin{itemize}
    \item High repetition: 18.01\% of the dataset contains the word 'asteroidi' (asteroid), indicating a significant bias towards astronomical content.
    \item Duplication: After removing duplicates and cleaning, only 17,103 unique entries remained, representing just 45\% of the original dataset.
    \item Quality issues: We found formatting errors, inappropriate content, and entries in non-Yoruba languages.
\end{itemize}

These findings highlight the pressing need for more stringent data curation practices in low-resource language datasets. The recent study by \citet{hernandez2022scalinglawsinterpretabilitylearning} on scaling laws and the interpretability of learning from repeated data offers valuable insights that influenced our methodology. Their extensive research reveals that data repetition can severely hinder model performance, particularly by disrupting the balance between memorization and generalization. Additionally, repeated data can obstruct the development of "induction heads," which are vital for in-context learning in large language models. Most importantly, their work underscores the pivotal role that high-quality training data plays in the effectiveness of language models. These insights significantly shaped our careful approach to developing Yankari, underscoring the critical need for rigorous data curation, quality control, and the preservation of diversity in our dataset.

\subsection{Data Processing Pipeline}
Our data processing pipeline consisted of several key steps:

\subsubsection{HTML Parsing and Text Extraction}
We implemented a robust HTML parsing system using BeautifulSoup to extract clean text from web pages. This process involved:
\begin{itemize}
    \item Removing all script and style elements
    \item Extracting text from relevant HTML tags (e.g., \texttt{<p>}, \texttt{<h1>}, \texttt{<h2>}, etc.)
    \item Preserving the document structure by maintaining paragraph boundaries
\end{itemize}
It is important to note that, beyond basic structural parsing, no specific sentence-level filtering based on the presence or absence of terminal punctuation was applied at this stage, as the primary goal was to capture full textual content from the identified relevant HTML elements.

\subsubsection{Deduplication}
We employed a two-step deduplication process:
\begin{enumerate}
    \item Exact matching at the document level to remove duplicate web pages
    \item Near-duplicate detection at the paragraph level using MinHash and Locality-Sensitive Hashing (LSH) techniques
\end{enumerate}

\subsection{Corpus Statistics}
Our final Yankari dataset consists of:
\begin{itemize}
    \item Total number of documents: 51,407
    \item Total number of tokens: 30,438,702
    \item Average tokens per document: 592.11
\end{itemize}

\subsection{Domain Analysis}
The distribution of web domains in our corpus reflects the diverse sources we targeted. The top 5 domains by number of documents are:
\begin{enumerate}
    \item yo.wikipedia.org: 32.70\%
    \item alaroye.org: 20.49\%
    \item www.bbc.com: 16.05\%
    \item www.awikonko.com.ng: 10.58\%
    \item yoruba.von.gov.ng: 4.94\%
\end{enumerate}

This distribution ensures a balance between encyclopedic content, news, and cultural discussions, providing a comprehensive representation of written Yoruba across various domains.

\subsection{Ethical Considerations and Excluded Content}
Throughout our data collection and processing, we prioritized ethical considerations:
\begin{itemize}
    \item We explicitly removed data from restricted sources, such as those with terms prohibiting use in NLP research.
    \item We filtered out suspected machine-translated content to maintain linguistic authenticity. This was primarily a manual process conducted by the author, a native Yoruba speaker, during data spot-checking and review. Documents exhibiting unnatural phrasing, common translation artifacts, or content inconsistent with known source characteristics were excluded.
    \item We removed inappropriate or offensive material to ensure the dataset's suitability for a wide range of applications. This was also a manual review process performed by the author. The guidelines focused on excluding hate speech, explicit adult content, and other materials generally considered unsuitable for a public research dataset.
    \item We respected copyright and intellectual property rights. Content was sourced primarily from publicly accessible websites. For sources like Wikipedia, explicit open licenses (e.g., CC-BY-SA) were followed. For other public news and blog content without explicit licenses for redistribution, we operated under the principles of fair use for non-commercial research, providing clear attribution via source URLs. Direct contact for explicit redistribution permission was not feasible for every source due to the scale of collection; this is a recognized challenge in web-scale data gathering and is noted as a limitation.
\end{itemize}

In line with our commitment to transparency, we acknowledge that our content filtering process may have introduced certain biases:
\begin{itemize}
    \item The removal of very short texts may have disproportionately affected certain types of content.
    \item Our focus on standard Yoruba may have led to the underrepresentation of regional dialects or colloquial expressions.
    \item Excluding content with non-Yoruba characters might have removed some culturally relevant content involving code-switching or borrowings.
\end{itemize}

\subsection{Output Format}
The final dataset is stored in JSONL format, with each line containing a separate JSON document with the following fields:
\begin{itemize}
    \item \textbf{text}: The main content of the document in Yoruba.
    \item \textbf{url}: The original URL from which the content was sourced.
    \item \textbf{source}: A code indicating the source of the document.
\end{itemize}

Here is a sample entry:

\begin{lstlisting}[breaklines=true]
{
  "text": "O ma se o! Ijamba oko ofurufu gba emi eeyan marun-un...",
  "url": "https://www.awikonko.com.ng/2024/03/o-ma-se-o-ijamba-oko-ofurufu-gba-emi.html",
  "source": "ACFLP"
}
\end{lstlisting}

\subsection{Quality Assurance}
To ensure the highest quality of our dataset:
\begin{itemize}
    \item We involved native Yoruba speakers in the data cleaning and validation process.
    \item We conducted regular spot checks throughout the data processing pipeline.
    \item We performed a final manual review of a randomly selected subset of the data to verify its quality and authenticity.
\end{itemize}

\subsection{Limitations and Potential Biases}
We acknowledge the following limitations and potential biases in our dataset:
\begin{itemize}
    \item Internet Bias
    \item Written Language Bias
    \item Source Bias
    \item Temporal Bias
    \item Standardization Bias
\end{itemize}

These limitations highlight areas for future work and expansion of the Yankari dataset.

\section{Limitations and Ethical Considerations}
While the Yankari dataset represents a significant contribution to Yoruba language resources for NLP, it is important to acknowledge its limitations and the ethical considerations that arise from its creation and potential use.

\subsection{Representation Bias}
\begin{itemize}
    \item Spoken Language: The dataset does not include samples of spoken Yoruba.
    \item Informal Variants: Internet sources may favor more formal language use.
    \item Demographic Skew: Internet access and content creation are not uniformly distributed across all Yoruba-speaking demographics.
\end{itemize}

\subsection{Diacritization Challenges}
\begin{itemize}
    \item Tonal Ambiguity: Incorrect or missing diacritical marks can cause ambiguity.
    \item Standardization Issues: The lack of a universally adopted standard for Yoruba orthography may result in inconsistencies.
\end{itemize}

\section{Conclusion}
The Yankari dataset represents a significant step forward in addressing the resource gap for Yoruba in Natural Language Processing. By providing a large-scale, high-quality, and ethically sourced corpus, we have laid a foundation for advancing NLP research and applications in this important West African language. Our rigorous methodology, which prioritizes data quality, diversity, and ethical considerations, sets a new standard for the development of language resources for low-resource languages.

The creation of Yankari highlights several critical challenges in developing NLP resources for languages like Yoruba, including the scarcity of diverse, high-quality online content, the complexities of automated processing for languages with limited existing NLP tools, and the ethical considerations surrounding data collection and potential misuse. By transparently discussing these challenges and our approaches to addressing them, we hope to contribute to the broader conversation on responsible AI development for diverse languages and cultures. Future work will focus on rigorously evaluating Yankari's utility in various downstream NLP tasks, such as language modeling, machine translation, and text classification, including comparative performance analyses against other available Yoruba corpora. Such evaluations will further quantify the benefits of Yankari's curated nature and diverse domain coverage.

The dataset is available on Hugging Face:
\href{https://huggingface.co/datasets/acflp/YANKARI}{https://huggingface.co/datasets/acflp/YANKARI}

\section*{Acknowledgments}
This work was performed independently. The author received no specific grant from any funding agency in the public, commercial, or not-for-profit sectors for the research described in this manuscript.

\bibliography{custom} 
\end{document}